\documentclass[journal]{IEEEtran}

\usepackage{amsmath,amsfonts,amssymb}
\usepackage{graphicx}
\usepackage{microtype}
\usepackage{booktabs}
\usepackage{siunitx}
\usepackage{url}
\usepackage{multirow}
\usepackage{makecell}
\usepackage[justification=centering]{caption}
\usepackage{tabularx}
\usepackage[table]{xcolor}
\usepackage{colortbl}
\usepackage{pifont}
\usepackage{arydshln}
\usepackage{cite}
\usepackage{stfloats}
\usepackage{hyperref}

\newcommand{\cmark}{\ding{51}}%
\newcommand{\xmark}{\ding{55}}%

\definecolor{mylightblue}{RGB}{220, 240, 255}
\definecolor{lightblue}{rgb}{0.88, 0.95, 1.0}

\begin{document}

\title{LAP: Fast \textbf{LA}tent Diffusion \textbf{P}lanner for Autonomous Driving}

\author{
Jinhao Zhang$^{*}$,
Wenlong Xia$^{*}$,
Zhexuan Zhou,
Haoming Song,
Youmin Gong,
Jie Mei$^{\dagger}$%
\thanks{$^{*}$Equal contribution.}
\thanks{$^{\dagger}$Corresponding author.}
}


\maketitle

\begin{abstract}
    Diffusion models have demonstrated strong capabilities for modeling human-like driving behaviors in autonomous driving, but their iterative sampling process induces substantial latency, and operating directly on raw trajectory points forces the model to spend capacity on low‑level kinematics, rather than high‑level multi-modal semantics. To address these limitations, we propose \textbf{LA}tent \textbf{P}lanner (LAP), a framework that plans in a VAE-learned latent space that disentangles high-level intents from low-level kinematics, enabling our planner to capture rich, multi-modal driving strategies. To bridge the representational gap between the high-level semantic planning space and the vectorized scene context, we introduce an intermediate feature alignment mechanism that facilitates robust information fusion. Notably, LAP can produce high-quality plans in \textbf{one single denoising step}, substantially reducing computational overhead. Through extensive evaluations on the large-scale nuPlan benchmark, LAP achieves \textbf{state-of-the-art} closed-loop performance among learning-based planning methods, while demonstrating an inference speed-up of at most $\mathbf{10\times}$ over previous SOTA approaches. 
\end{abstract}

\section{Introduction}
The efficacy of modern autonomous driving systems hinges on robust motion planning, capable of navigating complex, interactive environments \cite{chen2024end}. A central challenge is handling the inherent uncertainty and behavioral multimodality of real-world traffic, where multiple distinct yet equally plausible maneuvers may be available \cite{yang2023uncertainties, xiao2020multimodal}. While early rule-based systems offered interpretability, their hand-crafted logic is brittle and fails to scale to the long-tail of open-world scenarios \cite{fan2018baidu, chen2024end}. Consequently, the field has shifted towards data-driven Imitation Learning (IL), which excels at capturing nuanced, human-like behaviors from large-scale datasets \cite{le2022survey, teng2022hierarchical}. However, the standard IL objective is notoriously susceptible to mode-averaging, where the model collapses multiple valid expert trajectories into a single, physically infeasible path, fundamentally failing to represent the multi-modal nature of human decision-making \cite{strohbeck2020multiple}.

To overcome this limitation, Denoising Diffusion Probabilistic Models (DDPMs) have emerged as a powerful tool for modeling complex, multi-modal distributions\cite{liao2025diffusiondrive, ho2020denoising}. However, existing approaches modeling directly to raw trajectory waypoints are both computationally inefficient and semantic misalignment. This mirrors the core challenge of early image synthesis: operating in a high-dimensional pixel space forces the model to expend vast capacity on reconstructing low-level details rather than capturing high-level semantics\cite{rombach2022high}. A raw trajectory is analogous, as its high dimensionality is dominated by predictable kinematic redundancies (e.g., continuity, velocity limits, etc.) rather than strategic content. Consequently, training in this "waypoint space"  pixel level forces the model to waste capacity on modeling basic physics, distracting from the critical task of capturing the multi-modal semantics of driving strategy.

To address this challenge, we propose \textbf{LA}tent \textbf{P}lanner (LAP), a framework that decomposes trajectory generation into two specialized stages: learning strategic semantics in a compact latent space and reconstructing high-fidelity dynamics. We first design a Variational Autoencoder (VAE) \cite{kingma2013auto} to learn a low-dimensional latent space that captures the strategic essence of trajectories while abstracting away kinematic details. A conditional transformer-based diffusion model is then trained to plan on this latent space to focus on modeling the multi-modal distribution of high-level driving policies. However, latent compression introduces a representation gap: planning is done in a high-level semantic manifold, while scene context remains low-level vectorized features\cite{gao2020vectornet}. This mismatch weakens geometric grounding and alignment with map constraints. To bridge this modality gap, we further propose a fine-grained feature alignment method that better fuses semantic plans with low-level scene representations, enabling the model’s intermediate features to more effectively encode trajectory–scene interaction information and thereby improving decision robustness.


This two-stage approach yields a dual advantage: it dramatically improves generation efficiency by confining the diffusion process to a low-dimensional latent space, while simultaneously ensuring both strategic diversity and high-fidelity, kinematically feasible outputs. Through extensive evaluations on the large-scale nuPlan benchmark\cite{caesar2021nuplan}, we demonstrate that LAP establishes a new state-of-the-art in closed-loop performance among learning-based methods. Notably, this is accomplished with up to a 10$\times$ speed-up in inference, producing high-quality plans that substantially reduce the reliance on hand-crafted post-processing.

In summary, our key contributions are:

\begin{itemize}
\item We propose {\bf a latent diffusion framework} for autonomous planning that disentangles high-level strategic semantics from low-level kinematic execution, leading to improvements in both performance and computational efficiency.
\item We design {\bf a specialized trajectory VAE} that learns a compact, semantically rich latent space while ensuring high-fidelity, kinematically feasible reconstructions.
\item We introduce {\bf a novel intermediate feature alignment method} to bridge the gap between the high-level semantic planning space and low-level vectorized scene perception, facilitating the information interaction between them.
\item Our model, LAP, establishes a new state-of-the-art in closed-loop performance on the nuPlan benchmark for learning-based planners while demonstrating a substantial reduction in inference latency.
\end{itemize}

\section{Related Works}
\label{gen_inst}

Traditional motion planners are typically formulated as rule-based finite state machines with if-then-else logic\cite{zhou2022theories}, a paradigm valued for its interpretability and verifiable safety guarantees\cite{fan2018baidu, urmson2008autonomous}. However, the reliance on hand-crafted logic makes these systems inherently brittle and difficult to scale, as preventing conflicts between an ever-expanding set of rules becomes exponentially challenging\cite{grigorescu2020survey}. This fundamental limitation makes it intractable to exhaustively cover the infinite long-tail of novel scenarios found in dense, dynamic traffic, ultimately hindering their real-world applicability\cite{karnchanachari2024towards, lu2024activead}.

To overcome the scalability issues of rule-based planners, Imitation Learning (IL) has become the predominant paradigm, training policies to mimic the behavioral patterns of human experts from large-scale datasets\cite{le2022survey, codevilla2018end}. Such data-driven approach has spurred the development of increasingly sophisticated architectures, from early RNN-based models to powerful Transformer-based frameworks that capture complex, multi-modal environmental context\cite{bansal2018chauffeurnet, chitta2022transfuser, cheng2024pluto}. However, the standard IL objective is notoriously susceptible to mode-averaging, where multiple valid expert maneuvers are collapsed into a single and often kinematically infeasible trajectory\cite{strohbeck2020multiple}. This failure to capture the multi-modal nature of human decision-making motivates the shift towards generative models that can explicitly model diverse behavioral distributions.

To address the problem of mode-averaging, generative models have become an active area of research. An early, conceptually powerful approach was to perform planning in a latent space learned by a Variational Autoencoder, which could capture the semantic diversity of driving maneuvers\cite{zheng2024genad}. However, these methods were often constrained by the VAE's limited generative fidelity, struggling to produce kinematically realistic trajectories.

To resolve the issue of generation quality, recent works have adopted Denoising Diffusion Models as the core of the planning policy\cite{chi2023diffusion}. This paradigm excels at producing high-fidelity, diverse trajectories, with state-of-the-art frameworks demonstrating the ability to generate high-quality plans directly, removing the need for rule-based post-processing common in prior methods\cite{zheng2025diffusion}. Subsequent efforts have further improved sample diversity and efficiency\cite{jiang2025transdiffuser,liao2025diffusiondrive}. Nevertheless, a fundamental limitation shared by all these methods is their operation directly on raw trajectory waypoints. This forces the model to expend significant capacity learning low-level kinematics, rather than focusing its power on the more critical task of high-level strategic decision-making.
\begin{figure*}[t!] 
    \centering 
    \includegraphics[width=0.8\textwidth]{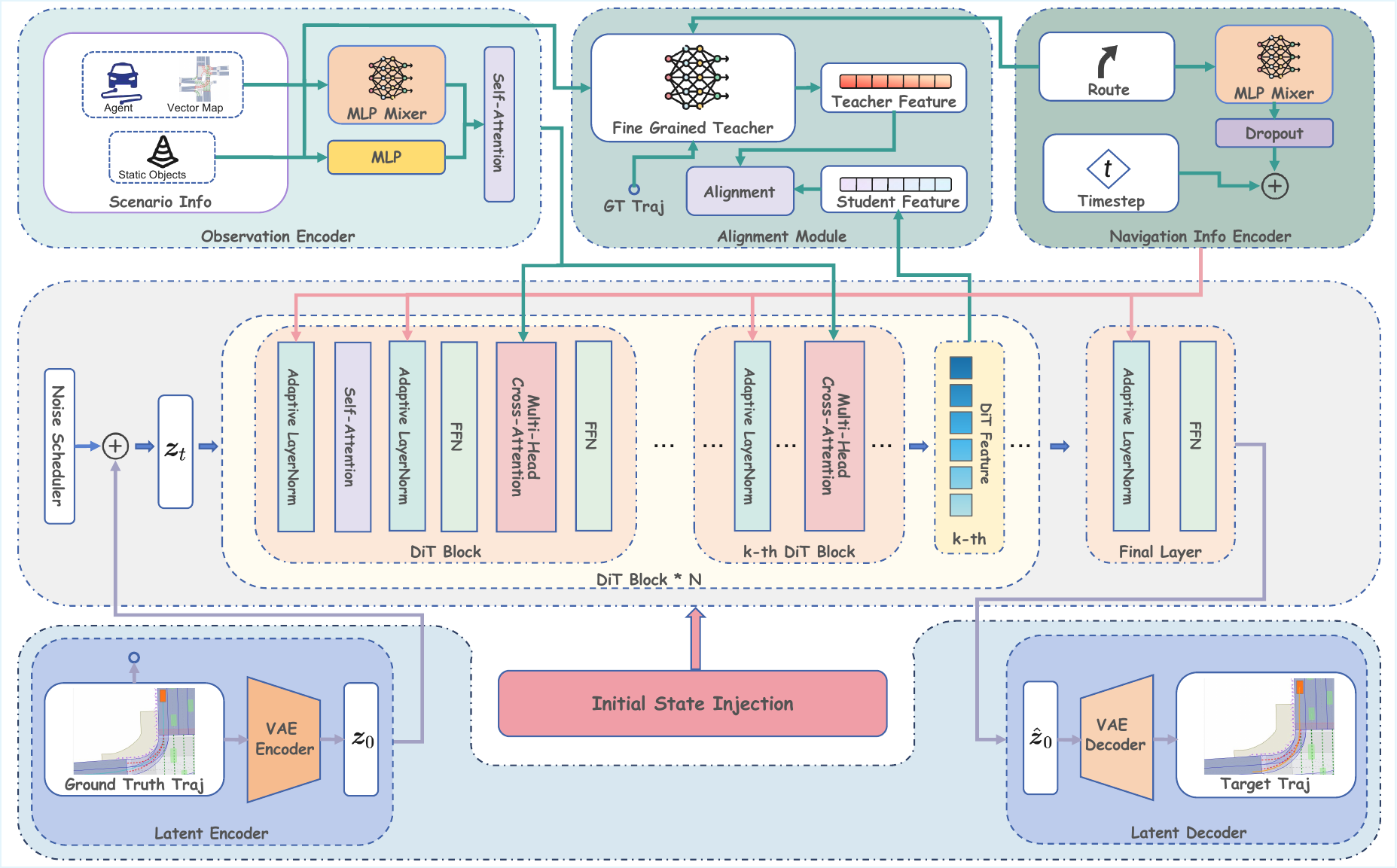} 
    \caption{Overall architecture of \emph{Latent Planner}} 
    \label{fig:overall_arch} 
\end{figure*}

\section{METHODOLOGY}
\label{method}


\subsection{Overview}
\textbf{Problem Formulation.}
Following the problem formulation of Diffusion Planner \cite{zheng2025diffusion}, our task is to plan the future trajectory for the ego-vehicle while jointly considering the potential future trajectories of $M$ neighbor agents, given a condition $\boldsymbol{C}$. This condition includes current vehicle states, historical data, lane information, and navigation information. The target is a collection of trajectories $\mathcal{T} = \{\tau_0, \tau_1, \dots, \tau_M\}$, where $\tau_0$ is the ego-vehicle's trajectory and $\{\tau_i\}_{i=1}^M$ are the trajectories of the neighbors. Each trajectory consists of states over a future horizon of $T$ timesteps:
$$
\mathcal{T} =
\begin{bmatrix}
    \tau_0 \\
    \tau_1 \\
    \vdots \\
    \tau_M
\end{bmatrix} = \begin{bmatrix}
    s_0^{(0)} & s_0^{(1)} & \cdots & s_0^{(T)} \\
    s_1^{(0)} & s_1^{(1)} & \cdots & s_1^{(T)} \\
    \vdots & \vdots & \ddots & \vdots\\
    s_M^{(0)} & s_M^{(1)} & \cdots & s_M^{(T)}
\end{bmatrix} \in \mathbb{R}^{(1+M) \times (1+T) \times 4},
$$
where $s_i^{(t)}$ denotes the state of agent $i$ at future timestep $t$. The state is represented by its coordinates and the sine and cosine of its heading angle.

\textbf{Overall Architecture.}
Our model, illustrated in Figure~\ref{fig:overall_arch}, is a Latent Diffusion Transformer \cite{peebles2023scalable} designed for trajectory planning. The training process consists of two stages:
First, we train a Variational Autoencoder \cite{kingma2013auto} responsible for encoding trajectories into a compact latent vector $\boldsymbol{z}_0$ and decoding it back to the original trajectory space. Second, we train a conditional diffusion model within the learned latent space. This model is trained to reverse a noising process, where it learns to denoise a noisy latent representation $\boldsymbol{z}_t$ back to the original latent representation $\boldsymbol{z}_0$, conditioned on the scene context $\boldsymbol{C}$. By operating in this compressed latent space, our model can efficiently generate multi-modal trajectory plans.
\vspace{-0.2cm}
\subsection{Trajectory Representation in latents}
Our proposed Trajectory Variational AutoEncoder ($\mathcal{V}$), comprising an encoder $\mathcal{E}$ and a decoder $\mathcal{D}$, is designed based on the Transformer architecture\cite{vaswani2017attention} to capture temporal dependencies within trajectory data. We augment the standard VAE objective with an auxiliary differential loss, which regularizes the model to produce smoother reconstructions. The encoder $\mathcal{E}$ compresses each trajectory $\tau_i$ into an $L$-dimensional latent space $\mathcal{Z}$, from which the decoder $\mathcal{D}$ reconstructs the trajectory.

More specifically, the encoder $\mathcal{E}$ first projects the input trajectories $\mathcal{T}$ into a sequence of high-dimensional embeddings. A set of learnable queries then interacts with these embeddings through a self-attention to aggregate the trajectories' essential features. These queries are then processed to produce the mean $\boldsymbol{\mu} \in \mathbb{R}^{(1+M) \times L}$ and diagonal covariance matrix $\boldsymbol{\Sigma} \in \mathbb{R}^{(1+M) \times L \times L}$ of the latent distribution. The decoder's operation begins with sampling a latent representation $\boldsymbol{z}_0 \sim \mathcal{N}(\boldsymbol{\mu}, \boldsymbol{\Sigma})=q(\boldsymbol{z}|\mathcal{T})$ using the reparameterization trick. It then utilizes a distinct set of way-point queries, which attend to the latent representation $\boldsymbol{z}_0$ via a multi-head cross attention. This process allows the way-point queries to gather the necessary information for reconstruction, after which they are projected to generate the reconstructed trajectories $\hat{\mathcal{T}}$.

To enforce a structured and semantically disentangled latent space, we adopt the $\beta$-VAE framework. This approach balances the compactness of the latent space with reconstruction quality by adding a weight, $\beta$, to the Kullback-Leibler (KL) divergence term of the original VAE training loss. We use the Mean Squared Error (MSE) as our reconstruction loss, which includes a trajectory reconstruction term and a differential reconstruction term. Let $\Delta$ denote the forward temporal difference operator, the final training objective for our VAE is formulated as:
\begin{equation}
    \mathcal{L}_{\text{vae}}=\|\mathcal{T}-\hat{\mathcal{T}}\|^2+\lambda \|\Delta\mathcal{T}-\Delta\hat{\mathcal{T}}\|^2 + \beta D_{\text{KL}}\left[q(\boldsymbol{z}|\mathcal{T})\|p(\boldsymbol{z})\right],
     \label{eq:vae_loss}
\end{equation}
where $\lambda$ is introduced to enhance the reconstruction quality, $p(\boldsymbol{z})=\mathcal{N}(\boldsymbol{0},\textbf{I})$. In our experiments, we set $\lambda = 0.01$. 

\subsection{Planning on latents}
After obtaining the pre-trained VAE, we train our planner, denoted as $\boldsymbol{z}_{\theta}(\boldsymbol{z}_{t},t,\boldsymbol{C})$, in the latent space. The training objective is to recover the original latents, $\boldsymbol{z}_0$, from the noisy latents\cite{ramesh2022hierarchical}:
\begin{equation}
    \mathcal{L}_{\text{diff}} = \mathbb{E}_{\mathcal{T}}\mathbb{E}_{\boldsymbol{z}_0 \sim q(\boldsymbol{z}|\mathcal{T}),\boldsymbol{z}_t \sim q_{t0}(\boldsymbol{z}_t|\boldsymbol{z}_0)}\left[\|\boldsymbol{z}_{\theta}(\boldsymbol{z}_{t},t,\boldsymbol{C})-\boldsymbol{z}_0\|^2\right].
\end{equation}
The details of the architecture are provided below.

\subsubsection{Latent Diffusion Planner}

We build our Latent Planner upon Diffusion Planner~\cite{zheng2025diffusion}. The process begins by encoding the various input modalities. Specifically, we use two separate MLP-Mixers\cite{tolstikhin2021mlp} to encode the historical information of neighboring vehicles $\boldsymbol{S}_{\text{neighbor}} \in \mathbb{R}^{A \times D_{\text{neighbor}}}$, and the lane segment information $\boldsymbol{S}_{\text{lane}} \in \mathbb{R}^{P \times D_{\text{lane}}}$, respectively. A standard MLP is used to encode the static obstacle information $\boldsymbol{S}_{\text{static}} \in \mathbb{R}^{D_{\text{static}}}$. Here, $A$ denotes the number of historical time steps, $P$ is the number of lane segments, and $D_{\text{neighbor}}$, $D_{\text{lane}}$, and $D_{\text{static}}$ are the respective feature dimensions. The resulting encoded features are then fed into a transformer encoder to obtain the fused feature representation $\boldsymbol{F}_{\text{scenario}}$.
Finally, to condition the trajectory generation on the environmental context, we fuse the encoded scene features $\boldsymbol{F}_{\text{scenario}}$ into the latent trajectory query $\boldsymbol{z}$ via a multi-head cross-attention mechanism:
\begin{equation}
    \boldsymbol{z} = \text{MHCA}\left(Q=\boldsymbol{z},K=V=\boldsymbol{F}_{\text{scenario}}\right).
\end{equation}
Additionally, we incorporate navigation information $\boldsymbol{S}_{\text{route}} \in \mathbb{R}^{(K \times P) \times D_{\text{route}}}$, where $K$ is the number of route lanes and $D_{\text{route}}$ is its feature dimension. We use MLP-Mixer to encode this information, and the resulting embedding, along with the diffusion timestep $t$, is injected into the model via adaptive layer normalization.

\subsubsection{Initial State Injection}

\begin{figure}[t]
\centering
\includegraphics[width=0.75\columnwidth]{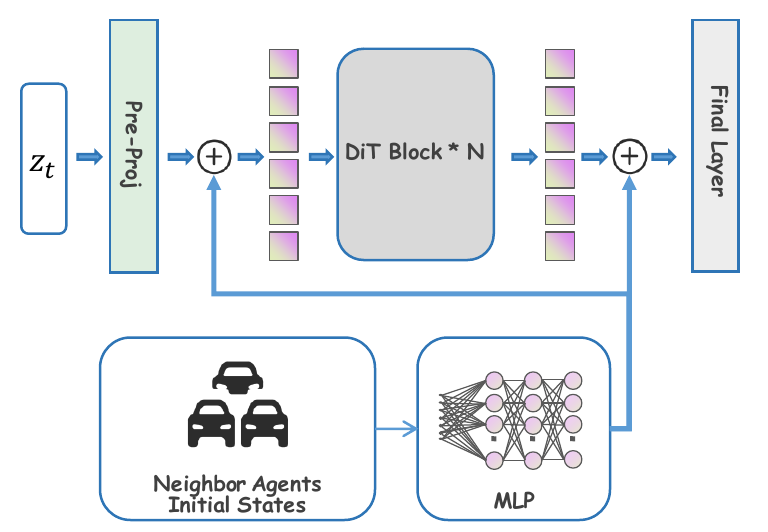}
\caption{Initial State Injection.}
\label{fig:isi}
\end{figure}

Modeling the future trajectories of surrounding vehicles is inherently challenging due to the lack of explicit knowledge regarding their instantaneous kinematic states. We observe that without explicit conditioning on these initial states, the prediction model struggles to converge. To address this, we inject the initial state of the surrounding vehicles, $s_{\text{init}}=[s_1^{(0)},\dots,s_M^{(0)}]$, as a conditional prior into the denoising process (Figure~\ref{fig:isi}). Specifically, we encode $s_{\text{init}}$ using an MLP and add the resulting embedding $e_{\text{init}}$ to the input of the first DiT block and the output of the final block. This conditioning effectively anchors the prediction to a known starting point, significantly stabilizing the training process and improving convergence.

\subsubsection{Fine-grained Feature Alignment} 

Our core insight is inspired by REPA\cite{yu2024representation}: for transformer-based diffusion planners, the quality of intermediate representations is a bottleneck, since they must encode fine-grained trajectory–scene interactions where critical planning constraints are expressed, rather than only high-level intent. However, the modality gap between the planning space and the scene context makes effective fusion challenging, often yielding suboptimal intermediate features that fail to capture these interactions. To address this challenge, we introduce a fine-grained feature alignment method to enhance the interaction and fusion between the semantic planning space and the vectorized scene context, enabling the planner to learn intermediate representations that are more conducive to decision making. Compared to direct interaction between the high-level planning space and low-level vectorized scene context, information exchange within a unified, fine-grained vectorized space is significantly more straightforward\cite{gao2020vectornet}. Therefore, we begin by encoding fine-grained trajectory-scene interaction information in the vectorized space to serve as a target feature. This feature then acts as a guiding exemplar for the intermediate layers of our DiT to guide the trajectory–scene fusion process. Formally, let $f$ be the encoder for the trajectory $\mathcal{T}$ and the condition $\boldsymbol{C}$. Let $\boldsymbol{y}^*=f(\mathcal{T},\boldsymbol{C}) \in \mathbb{R}^{(1+M) \times D}$ be the target feature and $\boldsymbol{h}_k \in \mathbb{R}^{(1+M) \times D}$ be the intermediate feature from the $k$-th layer of DiT, where $D$ is the embedding dimension. We augment the training objective with the following additional regularization term:
\begin{equation}
    \mathcal{L}_{\text{alig}} = \mathbb{E}_{\mathcal{T}}\mathbb{E}_{\boldsymbol{z}_0 \sim q(\boldsymbol{z}|\mathcal{T}),t \sim \mathbb{U}(0,1),\boldsymbol{z}_t \sim q_{t0}(\boldsymbol{z}_t|\boldsymbol{z}_0)}\left[\text{sim}(h_{\phi}(\boldsymbol{h}_k),\boldsymbol{y}^*)\right], \label{eq:reg}
\end{equation}
where $\text{sim}(\cdot,\cdot)$ is a pre-defined similarity function and $h_{\phi}$ is a learnable MLP projection head. The training objective becomes:
\begin{equation}
    \mathcal{L}=\mathcal{L}_{\text{diff}} + \alpha \mathcal{L}_{\text{alig}},
\end{equation}
where $\alpha > 0$ is a hyperparameter that controls the tradeoff between denoising and alignment. Intuitively, 
the target features act as a geometric anchor that transfers the teacher model’s fine-grained understanding of trajectory–scene interactions to the latent planner, bridging the gap between abstract semantic planning and low-level physical constraints.


Building on the finding from \cite{agarwal2025towards} that pixel-level diffusion models are effective encoders for fine-grained features and condition-image alignment, we leverage the pixel-level predictor, Diffusion Planner\cite{zheng2025diffusion}, as the encoder $f$. The target feature $\boldsymbol{y}^*$ is then defined as the activations from its intermediate DiT block, which takes the clean trajectory $\mathcal{T}$ and the condition $\boldsymbol{C}$ as input. 

\subsubsection{Navigation Guidance Augmentation}

We identified a causal confusion effect in the reactive scenarios, where the planner learns spurious correlations between the ego-vehicle's actions and the reactive behaviors of surrounding agents, often at the expense of ignoring the underlying navigation commands. To mitigate this, one solution is to strengthen the guidance from the navigation information, effectively 'forcing' the ego-vehicle to adhere more closely to the navigation route. We employ classifier-free guidance\cite{ho2022classifier} to achieve this objective. Specifically, during training, we randomly drop out the navigation information with a probability $p$. This allows the model to learn both a conditional (with navigation) and an unconditional (without navigation) denoising process. At inference time, the denoising is performed using a linear interpolation of the outputs from these two modes:
\begin{equation}
    \tilde{\boldsymbol{z}}_{\theta}(\boldsymbol{z}_{t},t,\boldsymbol{S}_{\text{route}}) = (1 - \omega) \boldsymbol{z}_{\theta}(\boldsymbol{z}_{t},t,\varnothing) + \omega \boldsymbol{z}_{\theta}(\boldsymbol{z}_{t},t,\boldsymbol{S}_{\text{route}}), \label{eq:cfg}
\end{equation}
where $\omega$ is the guidance scale. Here, we omit the scene information $\boldsymbol{F}_{\text{scenario}}$ for simplicity. In practice, we find that $\omega=1$ works well for most scenarios, and thus set it as the default. Then, \eqref{eq:cfg} becomes:
\begin{equation}
    \tilde{\boldsymbol{z}}_{\theta}(\boldsymbol{z}_{t},t,\boldsymbol{S}_{\text{route}})=\boldsymbol{z}_{\theta}(\boldsymbol{z}_{t},t,\boldsymbol{S}_{\text{route}})
\end{equation}
This means that the forward pass of the unconditional model is no longer required, which further accelerates our model's inference speed.

\begin{table*}[t]
\centering
\caption{Closed-loop planning results on nuPlan dataset. For each group, the best and second-best results are highlighted in \textbf{bold} and \underline{underline}, respectively. *: Using pre-searched reference lines as model input provides prior knowledge, reducing the difficulty of planning compared to standard learning-based methods.}
\label{tab:total_planner_compare}
\resizebox{0.8\textwidth}{!}{%
\begin{tabular}{llccc|ccc}
\toprule
\multirow{2}{*}{\textbf{Type}} & \multirow{2}{*}{\textbf{Planner}} & \multicolumn{3}{c|}{\textbf{Non-Reactive}} & \multicolumn{3}{c}{\textbf{Reactive}} \\
\cmidrule(lr){3-5} \cmidrule(lr){6-8}
& & \textbf{Val14} & \textbf{Test14-hard} & \textbf{Test14} & \textbf{Val14} & \textbf{Test14-hard} & \textbf{Test14} \\
\midrule
Expert & Log-Replay & 93.53 & 85.96 & 94.03 & 80.32 & 68.80 & 75.86 \\
\midrule
\multirow{7}{*}{\makecell[l]{Rule-based \\ \& Hybrid}} & IDM & 75.60 & 56.15 & 70.39 & 77.33 & 62.26 & 74.42 \\
& PDM-Closed & 92.84 & 65.08 & 90.05 & 92.12 & 75.19 & 91.63 \\
& PDM-Hybrid & 92.77 & 65.99 & 90.10 & 92.11 & 76.07 & 91.28 \\
& PLUTO & 93.17 & \textbf{80.08} & 92.60 & 90.78 & 76.88 & 91.65 \\
& Diffusion Planner w/ refine. & 94.24 & \underline{78.91} & \underline{94.13} & \textbf{92.76} & \textbf{81.66} & 91.85 \\
\rowcolor{mylightblue} & LAP w/ refine. ($o1s1$,Ours) & \underline{94.7} & 78.26 & 93.9 & 91.85 & 78.97 & \textbf{92.35} \\
\rowcolor{mylightblue} & LAP w/ refine. ($o1s2$,Ours) & \textbf{95.03} & 78.06 & \textbf{94.4} & \underline{91.95} & \underline{79.89} & \underline{92.2} \\
\midrule
\multirow{6}{*}{Learning-based} & PDM-Open* & 53.53 & 33.51 & 52.81 & 54.24 & 35.83 & 57.23 \\
& UrbanDriver & 53.05 & 43.89 & 51.83 & 50.42 & 42.26 & 52.34 \\
& GC-PGP & 59.51 & 46.13 & 61.16 & 55.30 & 42.74 & 54.86 \\
& PLUTO w/o refine.* & 88.89 & 74.39 & \underline{90.11} & 78.92 & 59.30 & 80.37 \\
& Diffusion Planner & \textbf{89.64} & 75.44 & 88.84 & \textbf{82.80} & 68.95 & 82.61 \\
\rowcolor{mylightblue} & LAP ($o1s1$,Ours) & 89.15 & \underline{78.11} & 89.85 & 81.82 & \underline{69.05} & \underline{84.24} \\
\rowcolor{mylightblue} & LAP ($o1s2$,Ours) & \underline{89.37} & \textbf{78.52} & \textbf{90.64} & \underline{82.23} & \textbf{70.53} & \textbf{85.32} \\
\bottomrule
\end{tabular}
}
\end{table*}

\begin{figure*}[t]
    \centering
    \includegraphics[width=0.7\textwidth, trim={0cm 6.4cm 5cm 0cm}, clip]{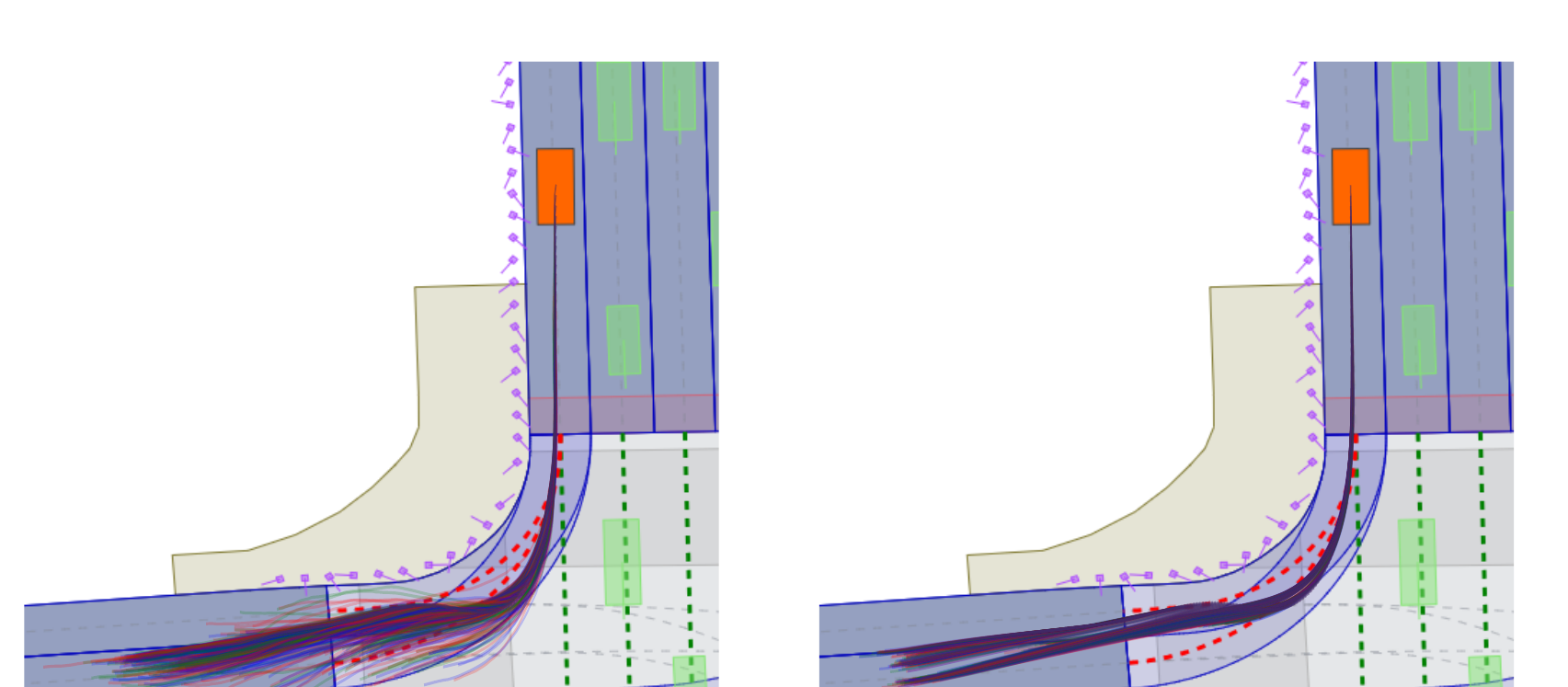}
    \caption{\textbf{Comparison of trajectory proposals for a right turn scenario.} 
    This figure illustrates the behavior of LAP (left) and Diffusion Planner (right) , which samples $K_{\text{mode}}=3$ candidate trajectories (thin colored lines) at each planning cycle. The proposals from LAP exhibit significant multi-modality, covering a diverse range of turning radii and speeds.}
    \label{fig:multitraj_turning}
\end{figure*}

\section{Experiments}
\label{sec:exp}

\textbf{Implementation Details.} 
We employ the same data augmentation techniques as Diffusion Planner\cite{zheng2025diffusion} to enhance the model's generalization to out-of-distribution (OOD) scenarios in closed-loop evaluations. Following previous work\cite{huang2023gameformer}
, we apply z-normalization to the original trajectories. Additionally, we scale the target latents $\boldsymbol{z}_0$ by the inverse of their global standard deviation to stabilize the training process\cite{rombach2022high}. Unless otherwise specified, we adopt the $L_2$ distance as the default similarity metric $\text{sim}(\cdot,\cdot)$. At inference time, we employ the DPM-Solver\cite{lu2022dpm} for accelerated sampling. Benefiting from the compactness and smoothness of the latent space, our model can perform rapid inference within two steps.

\textbf{Evaluation Setup}. 
We train and evaluate our model, LAP, on the nuPlan benchmark\cite{caesar2021nuplan}. nuPlan is a large-scale closed-loop autonomous driving simulation framework built upon 1,300 hours of real-world driving logs, encompassing 75 distinct, pre-classified scenarios. Its simulation models traffic agents as either Non-Reactive (replaying logs) or Reactive (governed by an IDM policy\cite{treiber2000congested}). For the SOTA comparison, we use a larger model with 192 hidden dimensions and 6 attention heads. In the ablation studies, excluding the one on sampling methods, we validate our approach with a smaller model with 128 hidden dimensions and 4 attention heads. Throughout the experiments, $o$ signifies the solver order and $s$ indicates the denoising steps.

\textbf{Baselines}. 
The baseline methods are categorized into three groups\cite{dauner2023parting}: Rule-based, Learning-based, and Hybrid. Hybrid approaches typically augment learning-based outputs with a refinement stage. To ensure a fair comparison, we apply a common post-processing refinement module\cite{sun2025generalizing} to our model's output without parameter tuning. 

\begin{itemize}
\item \textbf{IDM}\cite{treiber2000congested}: A classic rule-based car-following planner implemented by nuPlan.
\item \textbf{PDM}\cite{dauner2023parting}: The nuPlan challenge winner, with rule-based (\textbf{PDM-Closed}), learning-based (\textbf{PDM-Open}), and hybrid (\textbf{PDM-Hybrid}) versions.
\item \textbf{GC-PGP}\cite{hallgarten2023prediction}: A learning-based predictive model designed for goal-oriented navigation on lane graphs.
\item \textbf{UrbanDriver}\cite{scheel2021urban}: A learning-based planner using policy gradient optimization.
\item \textbf{PLUTO}\cite{cheng2024pluto}: An enhanced version of \textbf{PDM-Open} that uses contrastive learning for scene understanding.
\item \textbf{Diffusion Planner}\cite{zheng2025diffusion}: A specifically designed transformer-based diffusion model for high performance motion planning.
\vspace{-0.1cm}
\end{itemize}

\textbf{Evaluation Metrics}.
The nuPlan benchmark is derived from the 14 scenario categories, with each category containing up to 100 distinct scenes. The nuPlan framework provides three primary evaluation scores: an Open-Loop Score (OLS), a Non-Reactive Closed-Loop Score (NR-score), and a Reactive Closed-Loop Score (R-score). Consistent with prior research indicating a weak correlation between open-loop prediction and closed-loop planning effectiveness\cite{chen2024end}, we focus exclusively on the closed-loop scores.

\begin{table}[t]
    \centering
    \caption{Inference latency test}
    \label{tab:planner_inference_time}
    \resizebox{\columnwidth}{!}{%
        \begin{tabular}{lcccc}
            \toprule
            Planner & Inference time (ms) $\downarrow$ & score $\uparrow$  & {GFLOPS} & {model size(M)}\\ 
            \midrule
            UrbanDriver& 67.47 &43.89 & {-} & {-}\\
            GC-PGP& 39.24 &46.13 & {-} &{-}\\
            Pluto  w/o refine & 745.87 &73.61 & {\textbf{0.994}} & {\textbf{4.24}} \\
            DiffusionPlanner& 202.60 &75.44 & {1.38} &{\underline{6.04}} \\
            Latent Planner($o1s1$)&\colorbox{lightblue}{\textbf{18.81}}& \colorbox{lightblue}{\underline{78.11}} & \colorbox{lightblue}{{\underline{1.30}}} & \colorbox{lightblue}{{7.03}}\\
            Latent Planner($o1s2$)&\colorbox{lightblue}{\underline{21.69}}& \colorbox{lightblue}{\textbf{78.52}} & \colorbox{lightblue}{{1.33}} & \colorbox{lightblue}{{7.03}}\\
            \bottomrule
        \end{tabular}%
    }
\end{table}

\subsection{Main Results}
The comparison results with state-of-the-art methods on the nuPlan dataset are presented in Table \ref{tab:total_planner_compare}. Compared to all learning-based baselines, LAP achieves SOTA performance on most benchmarks within just two denoising steps. Notably, on the challenging Test14 hard dataset, our method shows a significant performance advantage over other learning-based methods, indicating its capability to make better decisions in complex environments. With the addition of PDM-based post-processing module, LAP's performance improves on most benchmarks, becoming comparable to SOTA rule-based and hybrid methods, and even surpassing human performance. Interestingly, on the Test14-hard NR benchmark, the performance of LAP shows a slight degradation after post-refinement. This suggests that in complex environments, LAP's decisions may already be superior to the PDM scoring module. 

We also evaluated the inference latency of the SOTA learning-based methods, and the results are shown in Table \ref{tab:planner_inference_time}. Although LAP and Diffusion Planner\cite{zheng2025diffusion} have comparable GFLOPs (most GFLOPs are spent in the scene encoder), LAP benefits from few-step sampling, so most of its computation is executed in parallel on the GPU with greatly reduced serial steps, resulting in up to a $10\times$ improvement in inference speed.

Another key advantage of planning in the latent semantic space is the ability to capture diverse, high-level driving strategies. This is demonstrated in Figure \ref{fig:multitraj_turning}, which visualizes the multi-modal trajectory proposals generated by LAP and Diffusion Planner\cite{zheng2025diffusion} for a right-turn scenario.

\subsection{Latent Planning vs. Pixel-level Planning}
To make the advantages of planning in the latent space more intuitive, we summarize in Table \ref{tab:lvsp} a comparison of its performance on the challenging Test14-hard NR benchmark, its capacity to model multi-modal trajectories, and its inference latency. Benefiting from the VAE, which compresses raw trajectories into high-level semantic representations, planning in the latent space can handle challenging scenarios more effectively while requiring less inference time. At the same time, planning in the semantic space also enables the planner to better model multi-modal driving strategies.

\begin{table}[h]
\centering
\caption{\textbf{Latent Planning vs. Pixel-level Planning.} APD: Average Pairwise Distance, FPD: Final Pairwise Distance.
}
\label{tab:lvsp}
\begin{tabular}{l c c}
    \toprule
    Planner & Test14-hard(NR) & Inference time (ms) \\
    \midrule
    Diffusion Planner     &  75.44   &   202.60 \\
    Latent Planner     &  \textbf{78.52}     &  \textbf{21.69} \\
    \midrule
    Planner & APD(m) & FPD(m) \\
    \midrule
    Diffusion Planner     & 0.88 & 1.98\\
    Latent Planner     &  \textbf{2.03} & \textbf{4.55} \\
    \bottomrule
\end{tabular}
\end{table}

\subsection{Latent Space Captures High-Level Trajectory Semantics}

\begin{figure}[t]
\centering
\includegraphics[width=0.85\columnwidth]{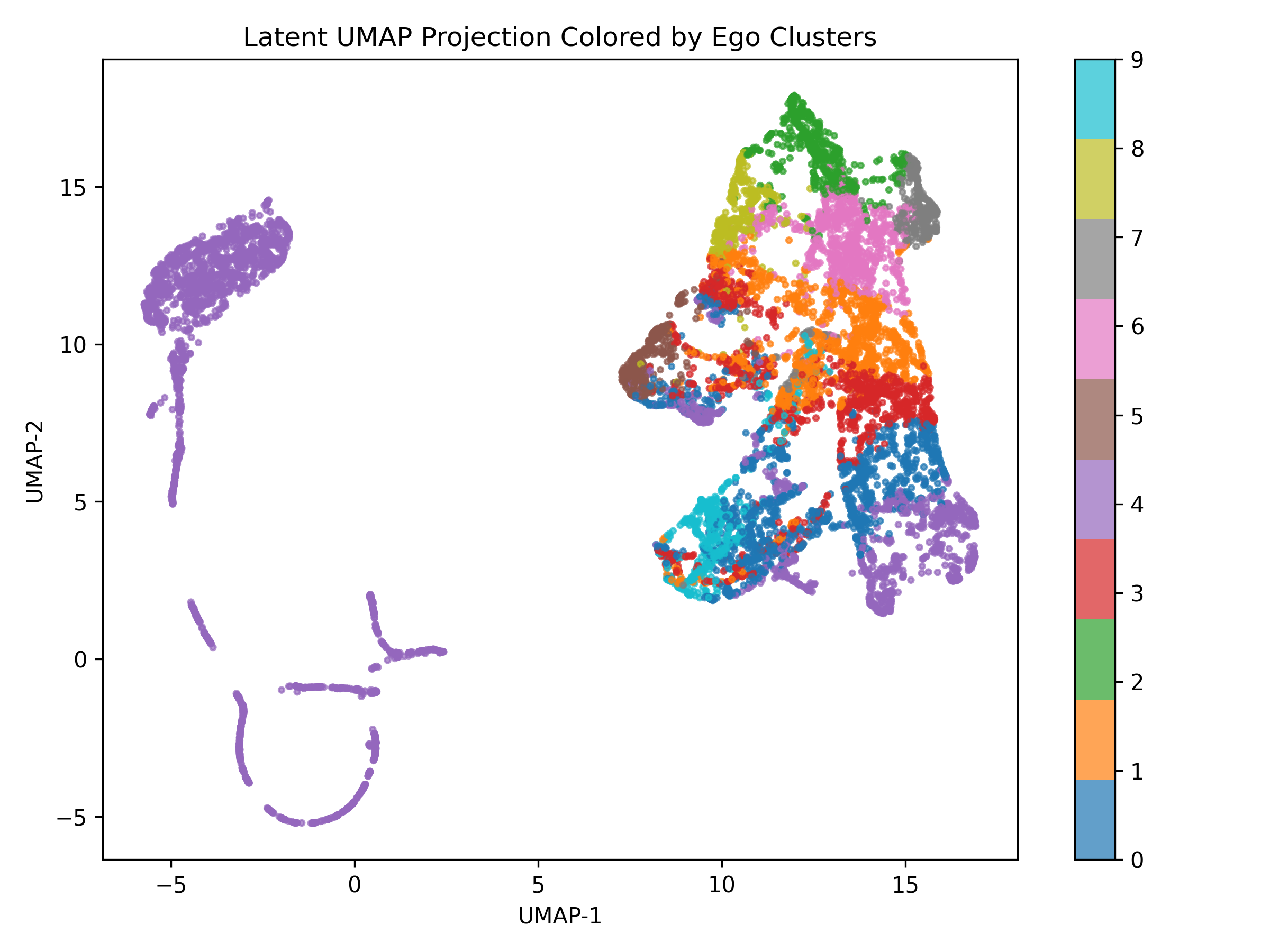}
\caption{UMAP visualization of the latent space.}
\label{fig:umap}
\end{figure}
To visually assess whether the latent space captures distinct intent categories of trajectories, we perform dimensionality reduction and color the latent codes according to their intent category. Since the trajectories do not come with explicit intent labels, we first sample 100k trajectories from the original dataset and run $k$-means($k=10$) clustering on them, using the resulting cluster assignments as pseudo intent labels. We then randomly sample another 100k trajectories, project their latent codes into 2D using UMAP\cite{mcinnes2018umap}, and color each point by its corresponding intent category, obtaining the visualization shown in Fig. \ref{fig:umap}. We observe that trajectories sharing the same intent form compact and mostly non-overlapping regions in the latent space, whereas different intents occupy well-separated areas. This demonstrates that the learned latent codes are well aligned with high-level semantic modes of driving. Additionally, we observe a subset of outliers (the purple region).

\begin{table}[t]
\centering
\caption{Impact of designed modules.}
\label{tab:module}
\begin{tabular*}{\columnwidth}{@{\extracolsep{\fill}}ccc|cc}
    \toprule
    \multicolumn{3}{c|}{ \textbf{Modules}} & \multicolumn{2}{c}{\shortstack{\textbf{Test14-hard} }} \\
    \cmidrule(lr){1-3} \cmidrule(lr){4-5}
    \textbf{ISI} & \textbf{Alig} & \textbf{CFG} &  \textbf{NR} & \textbf{R} \\
    \midrule
     \xmark   &   \xmark  & \xmark & 72.03 & 66.89 \\
    \cmark &  \xmark   & \xmark & 74.13 & 66.31 \\
    \cmark & \cmark & \xmark & \textbf{76.49} & 68.89 \\
    \cmark & \cmark & \cmark & 75.91 & \textbf{70.36} \\
    \bottomrule
\end{tabular*}
\end{table}

\begin{table}[t]
\centering
\caption{Impact of different sampling methods.}
\label{tab:sam}
\setlength{\tabcolsep}{3pt} 
\begin{tabular*}{\columnwidth}{c|@{\extracolsep{\fill}}ccc}
    \toprule
    \multirow{2}{*}{\textbf{\begin{tabular}[c]{@{}c@{}}Sampling\\ Method\end{tabular}}} & \multirow{2}{*}{\textbf{NFE}} & \multicolumn{2}{c}{\textbf{Test14-hard}} \\
    \cmidrule(lr){3-4}
     & & \textbf{NR} & \textbf{R} \\ 
    \midrule
     \cellcolor{red!15}$o1s1$ &  1   & 78.11 & 69.05 \\
     \cellcolor{red!15}$o1s2$  & 2   & 78.52 & 70.53    \\
     \cellcolor{red!15}$o1s3$  & 3   & 78.48 & 69.91    \\
     \cellcolor{red!15}$o1s10$  & 10  & 78.33 & 69.21    \\ \midrule
     \cellcolor{yellow!15}$o2s1$  & 2   & 78.17 & 69.93    \\
     \cellcolor{yellow!15}$o2s2$  & 4   & 78.23 & 70.45  \\
     \cellcolor{yellow!15}$o2s3$  & 6   & 77.46 & 69.53 \\
     \cellcolor{yellow!15}$o2s10$  & 20  & 77.13 & 69.19  \\
    \bottomrule
\end{tabular*}
\end{table}

\begin{table}[t]
\centering
\caption{\textbf{Analysis of alignment strategies} on Test14-hard dataset. Navigation guidance augmentation is not applied. 
}
\label{tab:distillation_analysis}
\resizebox{0.8\columnwidth}{!}{%
\begin{tabular}{ccccc}
    \toprule
    \multirow{2}{*}{\shortstack{\textbf{Alig.} \\ \textbf{Weight}}} & \multirow{2}{*}{\textbf{Objective}} & \multirow{2}{*}{\shortstack{\textbf{Stu.} \\ \textbf{Index}}} & \multicolumn{2}{c}{\textbf{Test14-hard}} \\
    \cmidrule(lr){4-5}
    & & & \textbf{NR} & \textbf{R} \\
    \midrule
    \rowcolor{gray!15}
     -     & - & -  & 74.13 & 66.31 \\
    \midrule
    \cellcolor{red!15}0.5      & $L_2$ & 3  & 76.38 & 68.28 \\
    \cellcolor{red!15}0.75     & $L_2$ & 3  & 76.49 & 68.89 \\
    \cellcolor{red!15}1        & $L_2$ & 3  & 76.65 & 68.91 \\
    \midrule
    \cellcolor{yellow!15}0.5  & $L_2$ & 2  & 75.23 & 67.48 \\
    \cellcolor{yellow!15}0.75 & $L_2$ & 2  & 75.32 & 70.21 \\
    \cellcolor{yellow!15}1    & $L_2$ & 2  & 75.06 & 66.72 \\
    \midrule
    0.5  & \cellcolor{green!15}$L_2$  & 3  & 76.38 & 68.28 \\
    0.5  & \cellcolor{green!15}cos.sim.     & 3  & 75.98 & 65.08 \\
    \midrule
    0.5  & $L_2$  & \cellcolor{blue!15}3   & 76.38 & 68.28 \\
    0.5  & $L_2$  & \cellcolor{blue!15}2   & 75.32 & 70.21 \\
    0.5  & $L_2$  & \cellcolor{blue!15}1   & 76.64 & 68.17 \\
    \bottomrule
\end{tabular}%
}
\end{table}

\subsection{ABLATION STUDIES}
In this subsection, we conduct a comprehensive ablation study of our method. The detailed results are presented below.

\textbf{Effect of Designed Modules.} 
Table \ref{tab:module} shows the effectiveness of our proposed modules, where ISI, Alig, and CFG denote \textit{Initial State Injection}, \textit{Fine-grained Feature Alignment}, and \textit{Navigation Guidance Augmentation}, respectively. By injecting the initial state of surrounding vehicles, the model's performance is enhanced in the Non-Reactive environment but shows slight degradation in the Reactive environment due to causal confusion. With the further addition of the alignment term, performance in both environments improves significantly. Finally, the integration of CFG greatly alleviates the causal confusion problem and boosts performance in the reactive environment, at the cost of a marginal performance drop in the Non-Reactive environment.

\textbf{Different Sampling Methods.} 
Table \ref{tab:sam} shows the impact of different sampling methods, where
NFE denotes \textit{Number of Function Evaluations}. Thanks to the latent space-based design, our model achieves excellent performance with just one denoising step, and two steps bring further improvement. However, performance degrades after three denoising steps. This is possibly due to excessively high decoding precision reducing the "flexibility" of the trajectories, which leads to a performance drop. This indicates that the latent space structure and the sampling method affect the model's performance jointly.

\textbf{Alignment Strategies.} 
Table \ref{tab:distillation_analysis} shows the impact of different alignment strategies, including varying the weights $\alpha$, similarity functions, and the DiT layer index $k$ from which the student features are obtained. As can be seen, taking features from the final layer of the DiT as students provides a more significant and stable performance improvement compared to intermediate layers. Furthermore, weights $\alpha$ within the $[0.5, 1]$ range are beneficial for model performance. Compared to the standard $L_2$ loss, relying solely on a cosine similarity constraint is insufficient and, in fact, leads to performance degradation in the Reactive environment.

\textbf{Impact of Different Teachers.} 
Table \ref{tab:dt} shows the impact on the model of using different alignment targets. We compare target features obtained from various intermediate layers of the Teacher DiT and those from different magnitudes of noise adding to the ground truth trajectory. Features from the Teacher DiT's first and last layers provide a more significant improvement than those from intermediate layers. Meanwhile, features obtained by encoding original trajectories with varying degrees of added noise consistently enhance model performance.

\textbf{Impact of Different VAEs.} 
Table \ref{tab:dv} shows the model performance with different VAEs, where $L$ denotes the latent dimension. It can be observed that the structure of the latent space is critically important for model performance. Although training the VAE for 200 epochs reduces its reconstruction error, it degrades the final performance of the model. This indicates that when planning in the latent space, a trade-off between its compactness and reconstruction fidelity is essential.


\begin{table}[h]
    \centering
    \scriptsize
    \setlength{\tabcolsep}{2.5pt}
    \renewcommand\arraystretch{1.1}

    \begin{minipage}[t]{0.49\columnwidth}
        \centering
        \resizebox{\linewidth}{!}{
        \begin{tabular}{l|cc}
            \toprule
            Teacher Feat. & \makecell{Test14-hard\\NR} & \makecell{Test14-hard\\R} \\ \midrule
            Layer3 & 76.38 & 68.28\\
            Layer2 & 74.05 & 65.92\\
            Layer1 & 76.63 & 67.67 \\
            Half Noise  & 76.61 & 66.80 \\
            Full Noise & 75.90 & 68.57\\
            \bottomrule
        \end{tabular}}
        \caption{\small \textbf{Impact of different targets.}}
        \label{tab:dt}
    \end{minipage}\hfill
    \begin{minipage}[t]{0.49\columnwidth}
        \centering
        \resizebox{\linewidth}{!}{
        \begin{tabular}{l|cc}
            \toprule
            VAE & \makecell{Test14-hard\\NR} & \makecell{Test14-hard\\R} \\ \midrule
            $\text{L}10\text{Epo}120$ & 76.38 & 68.28\\
            $\text{L}10\text{Epo}200$ & 74.72 & 66.69\\
            $\text{L}5\text{Epo}120$  & 74.41 & 63.40\\
            $\text{L}5\text{Epo}200$  & 72.76 & 63.00\\
            \bottomrule
        \end{tabular}}
        \caption{\small \textbf{Impact of different VAEs.}}
        \label{tab:dv}
    \end{minipage}

    \vspace{-0.3cm}
\end{table}

\subsection{ABLATION STUDIES}
\textbf{Training Details. }
We use the same training data as Diffusion Planner\cite{zheng2025diffusion}, which consists of 1 million scenarios randomly sampled from the nuPlan training data. The model was trained for 200 epochs on 4 NVIDIA RTX 5880 (48GB) GPUs with a batch size of 1024. We employed the AdamW optimizer\cite{loshchilov2017decoupled} with a learning rate of $5e^{-4}$, preceded by a 5-epoch linear warmup phase. A detailed summary of the setup is reported in Table \ref{tab:details}.

\textbf{Inference Details. }
We adopt a variance-preserving (VP) noise schedule:
\begin{equation}
    \sigma_t = (1-t)\beta_{\min}+t\beta_{\max}.
\end{equation}
We set the initial sampling temperature to 1.0 and used a default guidance scale of 1.0 throughout all experiments, with further hyperparameter details also provided in Table \ref{tab:details}.

\begin{table}[h] 
\centering 
\caption{Hyperparameters of LAP}
\label{tab:hyperparameters}
\begin{tabular}{@{}llll@{}}
\toprule
\textbf{Type}      & \textbf{Parameter}                     & \textbf{Symbol}                  & \textbf{Value} \\ \midrule
\multirow{5}{*}{VAE Training} 
                   & Num. encoder/decoder block           & -                                & 3              \\
                   & Dim. hidden layer                    & -                                & 128            \\
                   & Num. multi-head                      & -                                & 4               \\
                   & Regularization weight                    & $\beta$                                & $1e^{-6}$      \\
                   & Differential weight                    & $\lambda$                                & 0.01      \\
                    \midrule
\multirow{13}{*}{Planner Training} & Num. neighboring vehicles            & $M$                                & 10             \\
                   & Num. past timestamps                  & $A$                              & 21             \\
                   & Dim. neighboring vehicles            & $D_{\text{neighbor}}$            & 11             \\
                   & Num. lanes                           & -                                & 70             \\
                   & Num. points per polyline             & $P$                              & 20             \\
                   & Dim. lanes vehicles                  & $D_{\text{lane}}$                & 12             \\
                   & Num. navigation lanes                & $K$                              & 25             \\
                   & Num. encoder/decoder block           & -                                & 3              \\
                   & Uncond. probability                   & $p$                              & 0.1              \\
                   & Alig. weight                   & $\alpha$                              & 0.75              \\
                   & Student Feat. Idx                   & $k$                            & 3              \\
                   & Teacher Feat. Idx                 & -                              & 3              \\
                   & Dim. latent space                  & $L$                & 10             \\
                  \midrule
\multirow{4}{*}{Inference} & Noise schedule                       & -                                & Linear         \\
                   & Noise coefficient                    & $\beta_{\min}, \beta_{\max}$      & 0.1, 20.0      \\
                   & Temperature                          & -                                & 1.0            \\
                   & Guidance scale                          & $\omega$                                & 1.0            \\
                   \bottomrule
\end{tabular}  \label{tab:details}
\end{table}

\vspace{-14pt}
\section{CONCLUSION}
We introduce LAP, a latent diffusion framework that improves planning performance and efficiency by operating in a disentangled semantic space learned by a Trajectory VAE. By incorporating a novel fine-grained feature alignment method, LAP establishes a new state-of-the-art on the nuPlan benchmark while accelerating inference by up to $10\times$. 


%
%
\bibliographystyle{IEEEtran}
\bibliography{main}

\end{document}